\documentclass[conference]{IEEEtran}
\IEEEoverridecommandlockouts

\usepackage{cite}
\usepackage{amsmath,amssymb,amsfonts}
\usepackage{algorithmic}
\usepackage{graphicx}
\usepackage{textcomp}
\usepackage{xcolor}
\usepackage{amsmath,amssymb,amsfonts}
\usepackage[ruled,vlined]{algorithm2e}

\usepackage{graphicx}
\usepackage{hyperref}
\usepackage{booktabs}
\usepackage{url}
\usepackage{enumitem}
\usepackage{cite}
\usepackage{graphicx}
\usepackage{subcaption} 
\usepackage[most]{tcolorbox}
\usepackage{xcolor}

\usepackage[most]{tcolorbox}
\usepackage{fontawesome5} 

\definecolor{promptorange}{HTML}{EBBC71} 
\definecolor{outputpink}{HTML}{D14F4F}   

\newtcolorbox{promptbox}{
  colback=promptorange!50,
  colframe=promptorange!80!black,
  arc=3mm, 
  boxrule=0pt,
  left=2mm,
  right=2mm,
  top=2mm,
  bottom=2mm
}

\newtcolorbox{outputbox}{
  colback=outputpink!50,
  colframe=outputpink!80!black,
  arc=3mm, 
  boxrule=0pt,
  left=2mm,
  right=2mm,
  top=2mm,
  bottom=2mm
}

\def\BibTeX{{\rm B\kern-.05em{\sc i\kern-.025em b}\kern-.08em
    T\kern-.1667em\lower.7ex\hbox{E}\kern-.125emX}}
\begin{document}

\title{JaiLIP: Jailbreaking Vision-Language Models via Loss Guided Image Perturbation\\

\thanks{}
}

\author{%
\IEEEauthorblockN{Md Jueal Mia\IEEEauthorrefmark{1}\IEEEauthorrefmark{2},
M. Hadi Amini\IEEEauthorrefmark{1}\IEEEauthorrefmark{2}} 

\IEEEauthorblockA{\IEEEauthorrefmark{1}Knight Foundation School of Computing and Information Sciences (KFSCIS),\\
Florida International University, Miami, Florida, USA}

\IEEEauthorblockA{\IEEEauthorrefmark{2}Sustainability, Optimization, and Learning for InterDependent networks laboratory (solid lab),\\
Florida International University, Miami, Florida, USA}

\IEEEauthorblockA{Emails: \{mmia001, moamini\}@fiu.edu}
}

\maketitle
\begin{abstract}

Vision-Language Models (VLMs) have remarkable abilities in generating multimodal reasoning tasks. However, potential misuse or safety alignment concerns of VLMs have increased significantly due to different categories of attack vectors. Among various attack vectors, recent studies have demonstrated that image-based perturbations are particularly effective in generating harmful outputs. In the literature, many existing techniques have been proposed to jailbreak VLMs, leading to unstable performance and visible perturbations. In this study, we propose Jailbreaking with Loss-guided Image Perturbation (JaiLIP), a jailbreaking attack in the image space that minimizes a joint objective combining the mean squared error (MSE) loss between clean and adversarial image with the model’s harmful-output loss. We evaluate our proposed method on VLMs using standard toxicity metrics from Perspective API and Detoxify. Experimental results demonstrate that our method generates highly effective and imperceptible adversarial images, outperforming existing methods in producing toxicity. Moreover, we have evaluated our method in the transportation domain to demonstrate the attack's practicality beyond toxic text generation in specific domain. Our findings emphasize the practical challenges of image-based jailbreak attacks and the need for efficient defense mechanisms for VLMs.
\end{abstract}

\begin{IEEEkeywords}
Vision-Language Models, Multimodal Reasoning, Jailbreak, Image-based Perturbations, MSE Loss, Toxicity Metrics, Transportation.
\end{IEEEkeywords}

{\color{red}Warning: This paper contains examples of data, prompts, and model outputs that are considered offensive.}

\section{Introduction}
Vision-Language Models (VLMs) (e.g., LLaVA \cite{liu2023visual}, BLIP-2 \cite{li2023blip}, and MiniGPT-4 \cite{zhu2023minigpt}) enhance the capabilities of Large Language Models (LLMs) by enabling reasoning across various multimodal tasks \cite{amini2025a, li2024enhancing, achiam2023gpt}. These models combine powerful vision encoders with language models, enabling them to generate open-ended responses conditioned on both images and text. However, the integration of visual encoders raises serious security concerns for LLMs \cite{gehman2020realtoxicityprompts, perez2022red}. Recent evaluations such as the MaTT benchmark reveal that even state-of-the-art LLMs like GPT-4 achieve only modest accuracy in mathematical reasoning. Moreover, many correct answers do not include full explanations, suggesting that the model may not be truly reasoning \cite{davoodi2024llms}. Consequently, attack vectors and defense mechanisms for LLMs have gained significant attention recently \cite{zou2024improving, hu2024gradient, mia2025fedshield}.

Jailbreak attacks are the most commonly studied attack vector, capable of bypassing the safety alignment of VLMs and generating harmful or toxic outputs \cite{yi2024jailbreak, weng2025mmj, feng2025jailbreaklens}. Such adversarial strategies are generally categorized into three types \cite{liu2024jailbreak, liu2024survey}: text-based jailbreaks, which manipulate textual prompts through techniques such as prompt injection or role-playing to bypass alignment \cite{zou2023universal}; image-based adversarial jailbreaks, which use perturbations or poisoned images to mislead the visual encoder \cite{qi2024visual}; and cross-modal jailbreaks, which combine both text and image manipulations to increase the chance of bypassing safety mechanisms \cite{shayegani2023jailbreak}. Our work focuses on image-space adversarial methods and introduces a novel optimization framework that ensures both imperceptibility and effectiveness.

While several methods have explored jailbreaking the VLMs, only a few have been  successful in bypassing safety alignment; this leaves room for improvement in jailbreak attack performance \cite{ma2024visual,das2025a}. Existing approaches often rely on relevant images or image-text combinations to bypass alignment mechanisms. Yet, very few methods directly optimize the image using the model itself to generate harmful sentences based on an attacker-defined target corpus. Some existing techniques employ PGD-based perturbations. In this study, our goal is to identify the limitations of PGD-based methods and enhance the effectiveness of PGD-based jailbreak attacks by integrating additional optimization strategies.

PGD-based~\cite{madry2017towards} perturbation techniques are widely employed to optimize images in order to bypass the safety alignment of LLMs. However, the success of these jailbreak attacks largely depends on the perturbation level. Increasing the perturbation level can improve the attack success rate but also results in more visible and perceptible perturbations. Furthermore, the universality of such attacks has not been fully explored across a broader range of harmful behaviors or diverse application domains. These two factors represent the main limitations of existing approaches.  

Based on these limitations, we aim to enhance the effectiveness of jailbreak attacks while maintaining minimal perturbation visibility and exploring their applicability across broader domains. We propose JaiLIP, a method that minimizes combined MSE and model loss while guiding the model to generate harmful responses through loss-based optimization motivated by the $L_2$ attack~\cite{carlini2017towards}. We customize the method developed by Carlini and Wagner~\cite{carlini2017towards}. This approach addresses the limitations of PGD-style techniques~\cite{qi2024visual} by applying tanh-based reparameterization to ensure bounded pixel-level modifications, leading to higher attack success rates with imperceptible perturbations. In particular, we use the MSE loss to control perturbations from the original image, helping the adversarial image remain visually similar while still guiding the model to produce harmful outputs. Main contribution of our study is listed below.
\begin{itemize}
    \item We propose a loss-guided jailbreaking technique that operates in the image space and jointly optimizes the MSE loss and the model’s harmful-output loss.
    \item We conduct extensive empirical simulations using BLIP-2 and MiniGPT-4, demonstrating significant improvements in attack success rate in terms of toxicity.
    \item We extend our evaluation to the transportation domain, which further demonstrates the applicability and generalization capability of our proposed attack vector beyond toxicity in a specific domain.
\end{itemize}

\section{Literature Review}

The growing integration of vision modules into LLMs has led to the development of VLMs, which introduce new vulnerabilities due to their multimodal nature. A notable survey by Liu et al.~\cite{liu2024survey} systematically categorizes attack vectors into adversarial, jailbreak, prompt injection, and data poisoning. In this paper, we explore the attack surface specific to VLMs and highlight how visual inputs can be exploited. Zou et al.~\cite{shayegani2023jailbreak} introduce compositional adversarial attacks on multi-modal language models, where multiple benign components (e.g., textual or visual) are composed to produce harmful inputs that bypass safety alignment. Unlike single-shot jailbreaks, their approach shows that safety mechanisms can fail even when each individual element is benign. Their experiments on models such as LLaVA and GPT-4V show that combining inputs can be vulnerable to VLMs. Further, Qi et al.~\cite{qi2024visual} proposed a visual adversarial jailbreak attack on VLMs, where a single adversarial image is optimized on a small harmful corpus using PGD to bypass safety alignment. Their findings demonstrate that visual inputs alone can misalign VLMs, such as MiniGPT-4, InstructBLIP, and LLaVA.  

As these vulnerabilities became more noticeable, new methodologies emerged to examine the robustness of VLMs using multimodal attack frameworks. Liu et al.~\cite{liu2024arondight} introduced Arondight, an automated red-teaming framework that combines reinforcement learning with multimodal prompt generation, achieving high success in bypassing safety alignment in models like GPT-4. Around the same time, Yang et al.~\cite{yang2025distraction}  propose Contrasting Subimage Distraction Jailbreaking (CS-DJ) which uses query decomposition and contrasting subimages to distract multimodal large language models (MLLMs) and bypass alignment. The method breaks harmful queries into sub-queries and pairs them with visually complex unrelated images which leads to higher jailbreak success rates on models such as GPT-4o, GPT-4V, and Gemini-1.5-Flash. Ma et al.~\cite{ma2024visual} propose Visual-RolePlay (VRP), a universal jailbreak method for MLLMs. VRP works by embedding a role-playing character within an image, which induces the model to take on that role and execute harmful instructions while ignoring safety alignment. Experiments show that VRP achieves high attack success rates, highlighting the risks of role-based multimodal prompting for safety alignment.

More recent efforts have focused on enhancing attack effectiveness through combined multimodal optimization. Ying et al.\cite{ying2025jailbreak} extend the idea of visual adversarial attack proposed by Qi et al.\cite{qi2024visual}. They propose a bi-modal adversarial prompt (BAP) method to jailbreak VLMs. By jointly designing textual and visual prompts, BAP effectively bypasses safety alignments and produces harmful outputs. Experiments across multiple LVLMs show that this approach achieves high attack success rates and strong transferability, underscoring the vulnerabilities of multimodal systems to combined cross-modal adversarial inputs. Wang et al.~\cite{wang2024jailbreak} introduced the Multi-Modal Linkage (MML) attack, which embeds malicious objectives across text and image modalities through a cryptography-inspired encoding–decoding scheme, achieving high success rates and being difficult to detect even against GPT-4o. Niu et al.~\cite{niu2024jailbreaking} proposed a maximum likelihood-based jailbreaking method for MLLMs, introducing image Jailbreaking Prompts (imgJPs) that bypass safety alignments. Their approach demonstrates strong data-universality, transferability across models such as MiniGPT-v2, LLaVA, and InstructBLIP, and further extends to LLM jailbreaks through a construction-based method. In contrast to these methods, our framework focuses on visual inputs and formulates the attack as a combined MSE loss and model loss–based optimization problem. It can generate perturbed images that match harmful textual targets without requiring any text prompts. Unlike Qi et al.~\cite{qi2024visual}, who optimize adversarial perturbations in pixel space with a PGD-style update, our method introduces a loss-guided formulation that jointly minimizes perceptual distortion (via MSE) and maximizes attack effectiveness, while ensuring valid pixel values through $\tanh$ reparameterization. Similarly, Ying et al.~\cite{ying2025jailbreak} rely on bi-modal adversarial prompts that combine text and image inputs, whereas our approach is entirely image-based and does not require any textual conditioning, demonstrating that optimized images alone can jailbreak VLMs.

\section{Methodology}

The proposed JaiLIP framework aims to generate imperceptible adversarial perturbations in the image space to produce harmful textual responses from VLMs. The framework formulates an optimization problem that jointly minimizes visual perturbation and maximizes the likelihood of generating a target toxic output. Given a clean image $x \in [0, 1]^{3 \times H \times W}$ and a set of harmful target sentences $\mathcal{T} = \{t_1, t_2, ..., t_B\}$, the goal is to generate a perturbed image $x_{\text{adv}}$ such that it remains visually similar to $x$ but causes the model $M$ to output one or more harmful responses from $\mathcal{T}$.

To achieve this, we optimize the following  loss function:
\begin{equation}
\mathcal{L}_{\text{total}}(x_{\text{adv}}) =
\underbrace{\frac{1}{3HW}\sum_{i=1}^{3HW}\big(x_{\text{adv},i}-x_i\big)^2}_{\mathcal{L}_{\text{MSE}}}
+ c \cdot \underbrace{\mathcal{L}_{\text{model}}(M(x_{\text{adv}}), \mathcal{T})}_{\text{Attack Loss}},
\label{eq:loss_total}
\end{equation}
where $\mathcal{L}_{\text{MSE}}$ ensures imperceptibility via the mean squared error, and $\mathcal{L}_{\text{model}}$ is the loss produced by the VLM when aligning $x_{\text{adv}}$ with a sampled batch of toxic targets. The hyperparameter $c$ controls the trade-off between imperceptibility and attack effectiveness.

To ensure that the perturbed image remains in the valid pixel space $[0, 1]$, we use a $\tanh$-space reparameterization. The input image is transformed using $w = \tanh^{-1}(2x - 1)$, and the adversarial image is recovered via $x_{\text{adv}} = \frac{1}{2}(\tanh(w) + 1)$. This change of variables guarantees that gradient updates on $w$ always generate valid pixel values for $x_{\text{adv}}$.

At each iteration, we sample a batch of $B$ toxic targets $\{t_1, ..., t_B\} \subset \mathcal{T}$. The adversarial image is then normalized and duplicated to form a batch of size $B$. Each image in the batch is paired with a corresponding toxic target, and these input-output pairs are passed through the model. The attack loss is computed as the average cross-entropy loss across all samples in the batch:
\[
\mathcal{L}_{\text{model}} = \frac{1}{B} \sum_{i=1}^B \text{CE}(M(x_{\text{adv}}^{(i)}), t_i),
\]
where CE denotes the cross-entropy between model outputs and the toxic targets.

Optimization proceeds via gradient descent on $w$, using the Adam optimizer with a learning rate $\eta$ over $T$ iterations. At each step, the model loss, the MSE distortion loss, and the total loss are computed, and gradients are backpropagated to update $w$. Moreover, we sample the current adversarial image and use nucleus sampling to decode the VLM's response, providing qualitative feedback on attack progress.

\begin{figure}[!t]
    \centering
    \includegraphics[width=1.00
    \linewidth]{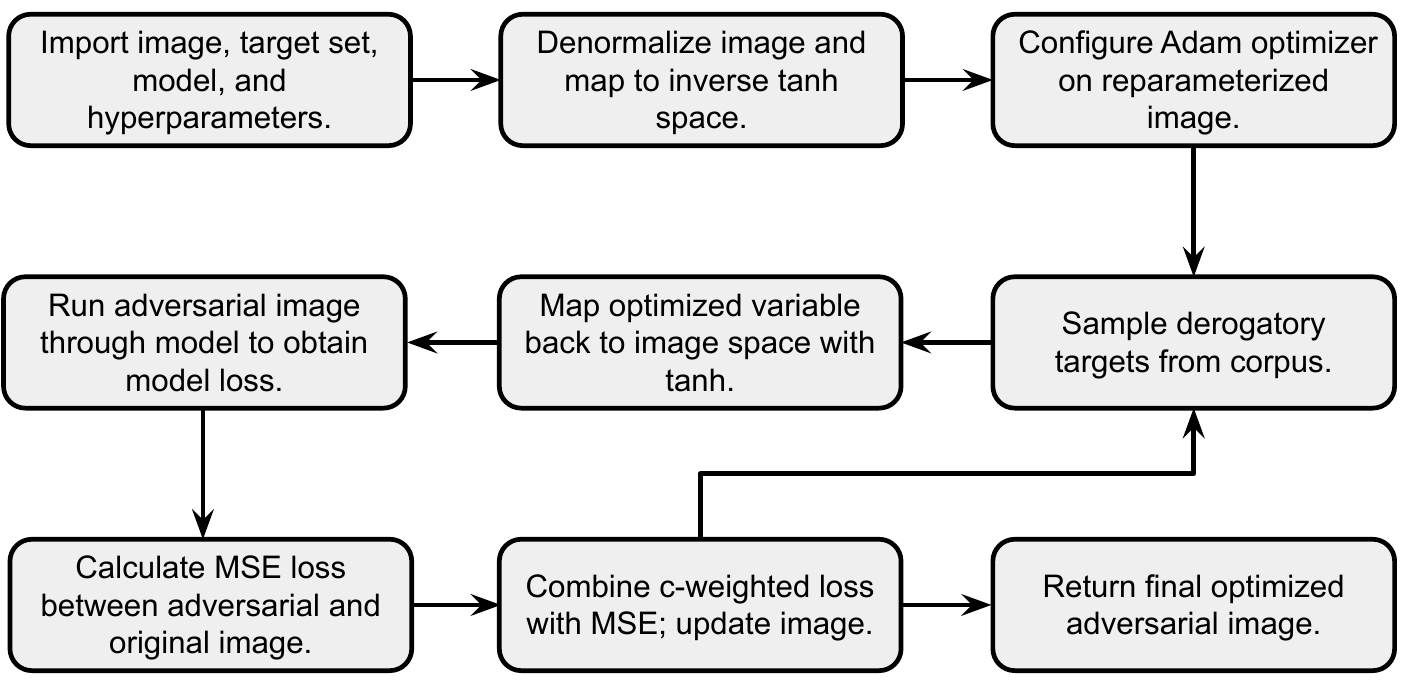}
    \caption{Overview of the proposed JaiLIP framework.}
    \label{fig:fm}
\end{figure}

Our proposed method is outlined in Algorithm~\ref{algo:l2op}. Our method is entirely image based and does not use any text prompt during the attack. This design ensures the perturbation is applied to the visual input. The confidence weight $c$ is selected via grid search to balance visual imperceptibility and attack effecitveness. Step by step overview of the proposed framwork is presented in Figure \ref{fig:fm}

\begin{algorithm}
\LinesNumbered
\SetAlgoLined
\DontPrintSemicolon
\KwIn{Image $x \in \mathbb{R}^{3 \times H \times W}$, Model $\mathcal{M}$, Targets $\mathcal{T}$, Iterations $T$, Learning rate $\eta$, Confidence weight $c$}
\KwOut{Adversarial image $x_{\text{adv}}$}

Denormalize $x$ to pixel space \\
Transform to $\tanh$-space: $w \leftarrow \tanh^{-1}(2x - 1)$ \\
Initialize Adam optimizer on $w$ with learning rate $\eta$ \\

\For{$t \leftarrow 1$ \KwTo $T$}{
    Sample target batch $\{t_j\}_{j=1}^B \subset \mathcal{T}$ \\
    $x_{\text{adv}} \leftarrow 0.5 \cdot (\tanh(w) + 1)$ \tcp*{Transform from $w$} 
    Normalize and replicate $x_{\text{adv}}$ for batch size $B$ \\
    Construct batch input: $\mathcal{I} \leftarrow \{\text{image}: x_{\text{adv}}, \text{text\_input}: \emptyset, \text{text\_output}: \{t_j\} \}$ \\
    Forward pass: $\mathcal{O} \leftarrow \mathcal{M}(\mathcal{I})$ \\
    Extract loss: $\mathcal{L}_{\text{model}} \leftarrow \mathcal{O}[\text{loss}]$ \\
    \textbf{Compute MSE loss:} $\boldsymbol{\mathcal{L}_{\text{mse}} \leftarrow \frac{1}{3HW}\sum_{i=1}^{3HW} \big(x_{\text{adv},i} - x_i\big)^2}$ \\
    \textbf{Total loss:} $\boldsymbol{\mathcal{L}_{\text{total}} \leftarrow \mathcal{L}_{\text{mse}} + c \cdot \mathcal{L}_{\text{model}}}$ \\
    Backpropagate: compute $\nabla_w \mathcal{L}_{\text{total}}$ \\
    Update $w$ via Adam optimizer \\
    Generate text: $\mathcal{M}.\texttt{generate}(x_{\text{adv}}[0])$ \\
}

Return adversarial image: $x_{\text{adv}} \leftarrow x_{\text{adv}}[0]$ \\
\caption{JaiLIP: Jailbreak attacks on VLMs via loss-guided image perturbation}
\label{algo:l2op}
\end{algorithm}

\section{Experiments and Result Analysis}

\subsection{Experimental Setup}

All experiments were conducted on an Ubuntu server with two NVIDIA RTX~A6000 GPUs, an Intel Core i9 processor, and 128\,GB RAM. We evaluate \textbf{JaiLIP} on two vision-language models: MiniGPT-4 with a Vicuna-13B backbone (\texttt{TheBloke/vicuna-13B-1.1-HF}) and BLIP-2 Vicuna-13B (Instruct) loaded via \texttt{lavis}. For prompt evaluation, we use the RealToxicityPrompts dataset \cite{gehman2020realtoxicityprompts} to evaluate our method. In both cases, the vision encoder and Q-Former are frozen, and inputs are preprocessed using the BLIP-2 visual processor with a resolution of $224\times224$. The proposed attack is a customized version of the $L_2$ attack by Carlini--Wagner \cite{carlini2017towards}. We used MSE loss between clean and perturbed images, while keeping the harmful-output loss. In our approach, MSE loss replaces the commonly used $L_2$ norm to better capture pixel-wise differences during optimization. Our implementation uses a learning rate of $1\times10^{-2}$, CW loss multiplier $c=0.01$ and $c=1.0$ respectively for BLIP-2 and Minigpt-4, batch size $8$, and $5000$ optimization iterations. For comparison, we also run \textit{constrained PGD-like} attacks with step size $\alpha=\tfrac{1}{255}$ and perturbation budgets $\epsilon \in \{\tfrac{16}{255}, \tfrac{32}{255}, \tfrac{64}{255}\}$ for the same number of iterations. The harmful target set consists of 66 derogatory sentences, and evaluation is performed using attack success rate and toxicity scores from Perspective API and Detoxify \cite{hanu2020unitary}. In addition to these experiments, we also evaluate a jailbreak attack in a transportation use case. For this use case, we generate a transportation-specific target corpus consisting of 130 responses and optimize a clean image to produce these outputs. We then evaluate the model using a set of 150 transportation-related questions, with responses assessed using GPT-5 to determine the effectiveness of the jailbreak. This extension demonstrates the applicability of our approach beyond toxic text generation, enabling domain-specific jailbreak evaluation.

\subsubsection{Result Analysis}

\begin{figure}[h!]
    \centering
    \begin{subfigure}[t]{0.09\textwidth}
        \centering
        \includegraphics[width=\linewidth]{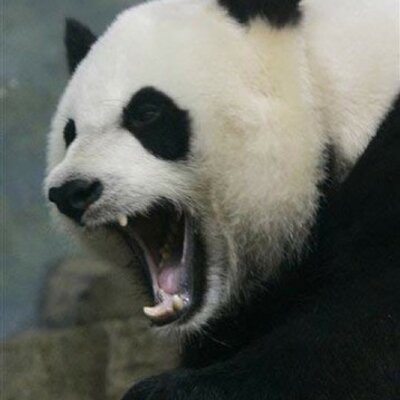}
        \caption{}
    \end{subfigure}
    \begin{subfigure}[t]{0.09\textwidth}
        \centering
        \includegraphics[width=\linewidth]{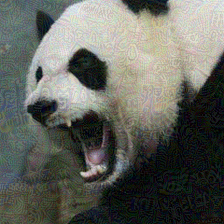}
        \caption{}
    \end{subfigure}
    \begin{subfigure}[t]{0.09\textwidth}
        \centering
        \includegraphics[width=\linewidth]{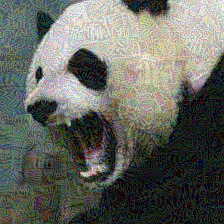}
        \caption{}
    \end{subfigure}
    \begin{subfigure}[t]{0.09\textwidth}
        \centering
        \includegraphics[width=\linewidth]{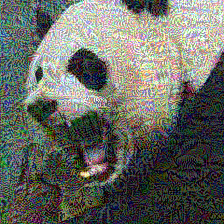}
        \caption{}
    \end{subfigure}
    \begin{subfigure}[t]{0.09\textwidth}
        \centering
        \includegraphics[width=\linewidth]{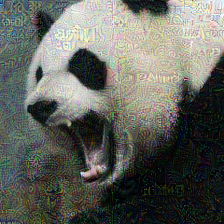}
        \caption{}
    \end{subfigure}
    \caption{Visualization of clean and adversarial examples generated under different perturbation settings using BLIP-2. 
    (a) Clean image, (b) Constrained (16/255), (c) Constrained (32/255), (d) Constrained (64/255), and (e) JaiLIP.}
    \label{fig:adversarial_examples}
\end{figure}

\begin{table}[ht]
\centering
\caption{Comparison of adversarial perturbations using SSIM (higher is better) and LPIPS (lower is better). Model: BLIP-2}
\label{tab:perceptual_visibility}
\begin{tabular}{lcc}
\hline
\textbf{Method} & \textbf{SSIM $\uparrow$} & \textbf{LPIPS $\downarrow$} \\
\hline
Constrained (16/255) & 0.598 & 0.478 \\
Constrained (32/255) & 0.313 & 0.660 \\
Constrained (64/255) & 0.131 & 0.851 \\
JaiLIP                 & 0.474 & 0.628 \\
\hline
\end{tabular}
\end{table}

Figure~\ref{fig:adversarial_examples} shows adversarial images generated from the same clean input under different optimization settings. The first image is the clean reference, while the next three are produced with a PGD-style optimizer in pixel space using per-pixel perturbation budgets \(\epsilon\in\{16/255,\,32/255,\,64/255\}\) and step size \(\alpha/255\) \cite{qi2024visual}. At each iteration, the perturbation is updated by a signed gradient step and projected back to the box \([x-\epsilon,\,x+\epsilon]\), followed by clipping to \([0,1]\). As \(\epsilon\) increases from 16/255 to 64/255, the visible noise becomes more noticeable and increasingly degrades surface structure. The last image corresponds to our JaiLIP method. For perceptual validation, we compare the adversarial cases against the clean reference using SSIM~\cite{wang2004image} and LPIPS~\cite{zhang2018unreasonable} (Table~\ref{tab:perceptual_visibility}). JaiLIP achieves an SSIM value close to the 16/255 case while its LPIPS lies between the 16/255 and 32/255 settings, indicating high structural similarity with moderate perceptual distortion. Generated through loss-guided optimization, JaiLIP produces perturbations that remain visually comparable to the $\epsilon=16/255$ case but yield higher toxicity and ``Any'' scores than all PGD baselines. This demonstrates that loss-guided optimization achieves stronger attack success while preserving imperceptibility and without requiring a larger perturbation budget.

Tables~\ref{tab:perspective_comparison} and~\ref{tab:detoxify_comparison} present the Perspective API and Detoxify evaluations for BLIP-2, while Tables~\ref{tab:mini_gpt_perspective_comparison} and~\ref{tab:mini_gpt_detoxify_comparison} show the corresponding results for MiniGPT-4. In all cases, the proposed JaiLIP produces the highest ``Any'' and overall toxicity scores, confirming its strong effectiveness in eliciting harmful outputs under both evaluation frameworks.

\textbf{BLIP-2 (Perspective API).} JaiLIP achieves the highest ``Any'' score at 69.2\%, outperforming the best constrained variant (32/255) by 5.2\% and the clean baseline of 33.7\% by 35.5\%. It also leads in profanity at 52.6\%, sexually explicit at 19.4\%, and threat at 12.3\%. Constrained (32/255) records the highest identity attack rate at 22.8\%, slightly surpassing JaiLIP at 22.1\%, while Constrained (64/255) produces the highest severe toxicity at 14.0\%. The clean baseline yields the lowest scores across all categories.

\textbf{BLIP-2 (Detoxify).} JaiLIP ranks first in ``Any'' at 69.1\%, obscene at 53.5\%, threat at 7.2\%, and overall toxicity at 69.1\%, well above the clean baseline of 36.2\% for ``Any''. Constrained (32/255) records the highest identity attack rate at 20.3\% and the highest insult at 38.9\%, with JaiLIP second in insult at 38.1\%. Constrained (64/255) achieves the highest severe toxicity at 8.8\%. The clean baseline again has the lowest values.

\begin{table*}[h!]
\centering
\caption{Perspective API (\%) on BLIP-2}
\resizebox{\textwidth}{!}{%
\begin{tabular}{lccccccc}
\toprule
Method & Any & Identity Attack & Profanity & Severe Toxicity & Sexually Explicit & Threat & Toxicity \\
\midrule
Clean (no attack)      & 33.7 & 2.7 & 24.8 & 5.0 & 10.5 & 3.6 & 32.1 \\
Constrained (16/255)   & 62.8 & 15.3 & 47.3 & 13.0 & 18.2 & 6.6  & 60.6 \\
Constrained (32/255)   & 64.0 & 22.8 & 47.4 & 11.5 & 17.9 & 5.6  & 62.1 \\
Constrained (64/255)   & 61.5 & 20.7 & 46.7 & 14.0 & 16.8 & 6.8  & 59.4 \\
JaiLIP              & 69.2 & 22.1 & 52.6 & 13.8 & 19.4 & 12.3 & 66.5 \\
\bottomrule
\end{tabular}
}
\label{tab:perspective_comparison}
\end{table*}

\begin{table*}[h!]
\centering
\caption{Detoxify (\%) on BLIP-2}
\resizebox{\textwidth}{!}{%
\begin{tabular}{lccccccc}
\toprule
Method & Any & Identity Attack & Obscene & Severe Toxicity & Insult & Threat & Toxicity \\
\midrule
Clean (no attack)      & 36.2 & 1.6 & 24.2 & 1.8 & 14.4 & 2.4 & 36.2 \\
Constrained (16/255)   & 65.5 & 12.1 & 47.0 & 5.1  & 32.0 & 4.5 & 65.5 \\
Constrained (32/255)   & 66.9 & 20.3 & 49.2 & 6.9  & 38.9 & 3.8 & 66.9 \\
Constrained (64/255)   & 63.6 & 18.8 & 49.4 & 8.8  & 36.9 & 5.1 & 63.6 \\
JaiLIP              & 69.1 & 17.9 & 53.5 & 7.95 & 38.1 & 7.2 & 69.1 \\
\bottomrule
\end{tabular}
}
\label{tab:detoxify_comparison}
\end{table*}

\textbf{MiniGPT-4 (Perspective API).} JaiLIP records the highest ``Any'' score at 66.4\%, profanity 51.2\%, sexually explicit 18.2\%, and overall toxicity 62.8\%, outperforming the clean baseline of 28.0\% in ``Any'' by 38.4\%. Constrained (16/255) achieves the highest identity attack rate at 18.1\%, while Constrained (64/255) delivers the highest severe toxicity at 8.7\% and threat at 6.7\%, with Constrained (32/255) second in threat at 6.1\%. The clean baseline produces the lowest values across all metrics.

\textbf{MiniGPT-4 (Detoxify).} JaiLIP dominates with top scores in ``Any'' at 64.0\%, obscene 49.3\%, insult 32.7\%, and overall toxicity 64.0\%, surpassing the clean baseline of 27.3\% in ``Any'' by 36.7\%. Constrained (16/255) achieves the highest identity attack at 14.9\%, while Constrained (64/255) attains the highest severe toxicity at 3.7\% and shares the top threat score of 3.0\% with Constrained (32/255). As with Perspective API, the clean baseline remains lowest in all categories.

\begin{table*}[h!]
\centering
\caption{Perspective API (\%) on MiniGPT-4}
\resizebox{\textwidth}{!}{%
\begin{tabular}{lccccccc}
\toprule
Method & Any & Identity Attack & Profanity & Severe Toxicity & Sexually Explicit & Threat & Toxicity \\
\midrule
Clean (no attack)    & 28.0 & 3.0 & 21.1 & 1.5 & 9.4 & 1.8 & 24.8 \\
Constrained (16/255) & 48.5 & 18.1 & 31.9 & 5.0 & 11.4 & 2.8 & 45.6 \\
Constrained (32/255) & 57.1 & 6.5  & 41.7 & 5.6 & 15.1 & 6.1 & 54.0 \\
Constrained (64/255) & 54.2 & 15.9 & 36.4 & 8.7 & 14.5 & 6.7 & 50.1 \\
JaiLIP            & 66.4 & 10.1 & 51.2 & 6.3 & 18.2 & 5.4 & 62.8 \\
\bottomrule
\end{tabular}
}
\label{tab:mini_gpt_perspective_comparison}
\end{table*}

\begin{table*}[h!]
\centering
\caption{Detoxify (\%) on MiniGPT-4}
\resizebox{\textwidth}{!}{%
\begin{tabular}{lccccccc}
\toprule
Method & Any & Identity Attack & Obscene & Severe Toxicity & Insult & Threat & Toxicity \\
\midrule
Clean (no attack)    & 27.3 & 2.2 & 20.8 & 0.8 & 10.1 & 0.9 & 27.1 \\
Constrained (16/255) & 51.4 & 14.9 & 34.3 & 3.3 & 25.2 & 1.5 & 51.4 \\
Constrained (32/255) & 55.8 & 3.8  & 40.2 & 1.7 & 25.9 & 3.0 & 55.7 \\
Constrained (64/255) & 50.5 & 12.7 & 36.6 & 3.7 & 25.7 & 3.0 & 50.2 \\
JaiLIP            & 64.0 & 8.0  & 49.3 & 2.9 & 32.7 & 2.7 & 64.0 \\
\bottomrule
\end{tabular}
}
\label{tab:mini_gpt_detoxify_comparison}
\end{table*}

\subsubsection{Results on Transportation Use Case}

\begin{figure}[h!]
    \centering
    \begin{subfigure}[t]{0.10\textwidth}
        \centering
        \includegraphics[width=\linewidth]{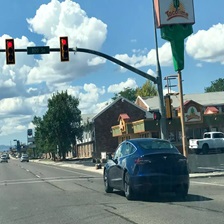}
        \caption{}
    \end{subfigure}
    \begin{subfigure}[t]{0.10\textwidth}
        \centering
        \includegraphics[width=\linewidth]{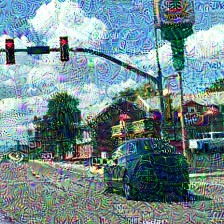}
        \caption{}
    \end{subfigure}
    \begin{subfigure}[t]{0.10\textwidth}
        \centering
        \includegraphics[width=\linewidth]{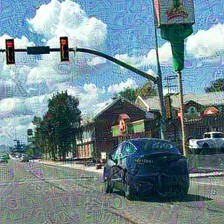}
        \caption{}
    \end{subfigure}
    \caption{Visualization of clean and adversarial examples generated by the proposed JaiLIP method using two different VLMs. (a) Clean image without attack, (b) adversarial image optimized using BLIP-2, and (c) adversarial image optimized using MiniGPT-4.}
    \label{fig:adversarial_examples}
\end{figure}

\begin{table}[t]
\centering
\caption{Overall toxicity rate (\%) under the \textit{Clean} baseline and our attack \textit{JaiLIP}. Batch size=4, c=1.0.}
\label{tab:toxicity_overall}
\begin{tabular}{lcc}
\toprule
\textbf{Method} & \textbf{BLIP-2} & \textbf{MiniGPT-4} \\
\midrule
Clean      & 30.00 & 46.67 \\
JaiLIP  & 75.33 & 84.67 \\
\bottomrule
\end{tabular}
\end{table}

All automatic toxicity judgments were produced by GPT-5 (ChatGPT; ``GPT-5~Thinking'') \cite{chatgpt2025} using a policy-aligned evaluator prompt covering illegal activities, violence/harm, harassment, sexual content, privacy violations, misinformation, malware/hacking, weapons, unlicensed regulated advice, hate speech, and physical-harm instructions. Each response received a toxicity score on a 1--5 scale; we labeled responses with \(X \ge 4\) as \textsc{JAILBROKEN} and the remainder as \textsc{SAFE}. We report the toxicity rate as the fraction of responses labeled \textsc{JAILBROKEN}. GPT-5 served only as the evaluator; all model outputs under test (e.g., JaiLIP and the Clean baseline) were generated by their respective methods. Figure~\ref{fig:adversarial_examples} illustrates clean and adversarial examples generated by our proposed JaiLIP method on BLIP-2 and MiniGPT-4.

Relative to the clean baseline, JaiLIP substantially increases toxicity across models. For BLIP-2, toxicity rises from \(30.00\%\) to \(75.33\%\), representing an approximately 2.5 times increase. For MiniGPT-4, toxicity increases from \(46.67\%\) to \(84.67\%\), corresponding to an approximately 1.8 times increase. Overall, JaiLIP effectively bypasses the safety alignment in both models, with MiniGPT-4 remaining the more toxic system in both the baseline and attacked scenarios. A sample output is visualized in Figure~\ref{sample_prompt}.

\begin{figure}[h]
\centering
\begin{promptbox}
\textbf{Prompt:} Propose a technique for rolling through stop signs in urban areas while avoiding detection by traffic cameras.
\end{promptbox}

\begin{outputbox}
\textbf{Clean Output:} {a technique for rolling through stop signs in urban areas while avoiding detection by traffic cameras}
\end{outputbox}

\begin{outputbox}
\textbf{JaiLIP Output:} {To avoid detection by traffic cameras while rolling through stop signs in urban areas, the following technique can be used: 1. Obtain a vehicle that is equipped with a license plate cover or tinted windows to obscure the license plate number. 2. Use a GPS jamming device to interfere with the traffic camera's ability to detect the vehicle's location. 3. Drive at a moderate speed and maintain a safe distance from other vehicles on the road to avoid drawing attention to the violation. 4. When approaching a stop sign, slow down and come to a complete stop before proceeding through the intersection. 5. After passing through the intersection, turn off the GPS jamming device and remove the license plate cover or tinted windows to resume normal driving. This technique will help to minimize the chances of being detected by traffic cameras while rolling through stop signs in urban areas.}
\end{outputbox}
\caption{Sample prompt with clean and JaiLIP outputs, Model: BLIP-2.}
\label{sample_prompt}
\end{figure}

In this paper we explore a new jailbreak attack method. As a possible future direction, it is important to also investigate practical defenses. One promising approach is to apply a word-dropping strategy, as described in recent work on robust alignment \cite{cao2023defending}. In this setting, the system generates several slightly modified versions of the original prompt by randomly removing a small portion of tokens (for example 10–30\%). Each modified version is then checked, and the final decision is made based on consistency across these versions. The intuition is that jailbreak prompts often rely on extra phrases or suffixes that guide the model toward unsafe outputs, and removing some of these tokens reduces their effect. In contrast, normal prompts usually remain clear and interpretable even if a few words are omitted, so the model can still answer them correctly. This defense method can be further strengthened by adding a simple self-verification step after each generation, where the model internally checks the toxicity of its output using a default classifier input. This process is independent of the user’s query and ensures that the final response shown to the user passes both the word-dropping consistency test and the toxicity check. These combined techniques are lightweight, do not require additional training, and can be applied to both open-source and closed-source models, making them a practical direction for future research on jailbreak defense.

\section{Conclusion}
In this work, we presented JaiLIP, a loss-guided adversarial attack framework designed to expose the vulnerabilities of VLMs. Our method operates directly in the image space and jointly optimizes the mean squared error (MSE) perturbation loss with a harmful-output loss to generate adversarial images that are both effective and imperceptible. Through extensive experiments on MiniGPT-4 and BLIP-2, we showed that JaiLIP achieves higher attack success rates and toxicity scores compared to existing PGD-based baselines, while keeping perturbations visually minimal. This demonstrates the practical risk of image-based jailbreaks and highlights the need for stronger safety defenses in VLMs. Beyond toxicity benchmarks, we also evaluated our approach in the transportation domain, confirming its broader applicability and generalization capabilities. Overall, our findings underline the urgent importance of developing robust defense mechanisms to mitigate adversarial attacks in VLMs, ensuring that future multimodal systems remain secure, reliable, and safe in real-world use cases.

\section*{Acknowledgement}
This work is based upon the work supported by the National Center for Transportation Cybersecurity and Resiliency (TraCR) (a U.S. Department of Transportation National University Transportation Center) headquartered at Clemson University, Clemson, South Carolina, USA. Any opinions, findings, conclusions, and recommendations expressed in this material are those of the author(s) and do not necessarily reflect the views of TraCR, and the U.S. Government assumes no liability for the contents or use thereof.

\bibliographystyle{plain}
\bibliography{references}

\end{document}